\pdfoutput=1

\documentclass[11pt]{article}

\usepackage[]{EMNLP2023}

\usepackage{times}
\usepackage{latexsym}

\usepackage[T1]{fontenc}

\usepackage[utf8]{inputenc}

\usepackage{microtype}

\usepackage{inconsolata}

\usepackage{amsfonts}
\usepackage{amsmath}
\usepackage{multirow}
\usepackage{adjustbox}
\usepackage{array}
\usepackage{makecell}

\title{Re-Temp: Relation-Aware Temporal Representation Learning \\for Temporal Knowledge Graph Completion}

\author{Kunze Wang$^1$ 
        \quad Soyeon Caren Han$^{1,2}$\thanks{\ \ Corresponding author. caren.han@sydney.edu.au} 
        \quad Josiah Poon$^1$\\ 
    $^1$The University of Sydney \\
    $^2$The University of Western Australia\\ 
    \tt kwan4418@uni.sydney.edu.au 
     \quad \{caren.han, josiah.poon\}@sydney.edu.au \\}

\begin{document}
\maketitle
\begin{abstract}
Temporal Knowledge Graph Completion (TKGC) under the extrapolation setting aims to predict the missing entity from a fact in the future, posing a challenge that aligns more closely with real-world prediction problems. Existing research mostly encodes entities and relations using sequential graph neural networks applied to recent snapshots. However, these approaches tend to overlook the ability to skip irrelevant snapshots according to entity-related relations in the query and disregard the importance of explicit temporal information. To address this, we propose our model, Re-Temp (Relation-Aware Temporal Representation Learning), which leverages explicit temporal embedding as input and incorporates skip information flow after each timestamp to skip unnecessary information for prediction. Additionally, we introduce a two-phase forward propagation method to prevent information leakage. Through the evaluation on six TKGC (extrapolation) datasets, we demonstrate that our model outperforms all eight recent state-of-the-art models by a significant margin.  
\end{abstract}

\section{Introduction}
A Knowledge Graph (KG) is a graph-structure database, composed of facts represented by triplets in the form of (\textit{Subject Entity, Relation, Object Entity}) such as (\textit{Alice, Is a Friend of, Bob}). In this graph, entities serve as nodes, and relations are depicted as direct edges connecting the nodes. However, facts in a KG are not static but undergo continuous updates over time. To incorporate temporal information into the KG, Temporal Knowledge Graphs (TKGs) are introduced. TKGs add the extra temporal information of each fact and extend each triple with a timestamp as a quadruplet (\textit{Subject Entity, Relation, Object Entity, Timestamp}). A TKG can be represented as a sequence of snapshots, where each snapshot represents a static knowledge graph for one specific timestamp.  

Temporal Knowledge Graph Completion (TKGC) aims to predict the missing entity from a query (\textit{Subject Entity, Relation, ?, Timestamp}) or (\textit{?, Relation, Object Entity, Timestamp}). TKGC is difficult and even large-scale pre-trained language models such as ChatGPT\cite{chatgpt} are prone to making factual errors\cite{borji2023categorical}. There are two main settings: interpolation and extrapolation setting. TKGC under the interpolation setting completes the facts in history, while TKGC under the extrapolation setting predicts facts at future timestamps. In this paper, we focus on TKGC in the extrapolation setting, which is more challenging and requires further improvement\cite{renet}. 

\begin{figure}
    \centering
    \includegraphics[width=1\linewidth]{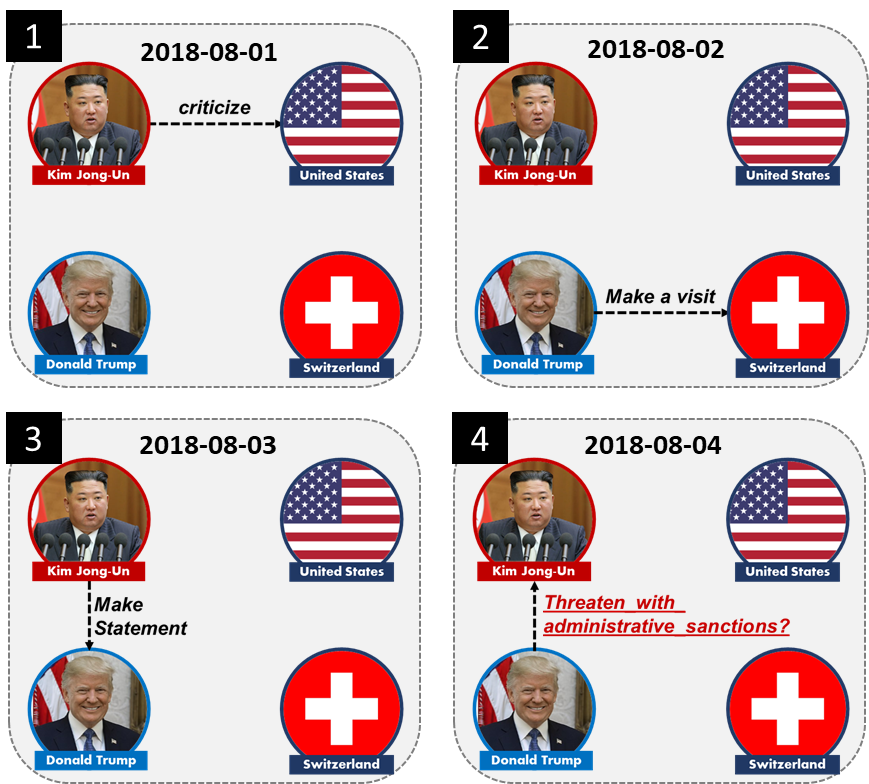}
    \caption{A case study of temporal knowledge graph completion under the extrapolation setting}
    \label{fig:case}
\end{figure}

Enormous attention has been focused on static KGC problems, and numerous models have been employed to encode entities and relations. However, a key question remains: how can a static KGC model be extended to incorporate temporal information for TKGC tasks? 
Recent works\cite{renet,regcn,cen,hismatch} have utilised sequential Graph Neural Networks (GNNs) to the previous snapshots for encoding the entities and relations. Then, they use a static score function as the decoder to assess the score of each candidate. Sequential GNNs are used because the facts shown in recent history can be helpful when making predictions in the future. An example is shown in Figure \ref{fig:case}, the previous facts (\textit{Kim Jong-Un, criticize, United States}) three days before and (\textit{Kim Jong-Un, Make Statement, Donald Trump}) one day before may imply (\textit{Donald Trump, Threaten with administrative sanction, Kim Jong-Un}) today. Since no explicit timestamp value is used, we can call it ``implicit temporal information''.  

However, to effectively encode the timestamp, the temporal information, two additional considerations arise: First, explicit temporal information is crucial. For instance, the validity score of (Donald Trump, Threaten with administrative sanction, Kim Jong-un) may differ between 2018 and 2023 as Donald Trump was the president in 2018 but not in 2023, affecting his ability to threaten another nation with administrative sanctions in 2023. The nature of entities can change over time, necessitating the consideration of explicit temporal information to encode time-dependent factors. Second, not all the facts in the recent history are relevant. Given historical facts (Kim Jong-un, criticize, United States, 2018-08-01),(Donald Trump, Make a visit, Switzerland, 2018-08-02) and (Kim Jong-un, Make Statement, Donald Trump, 2018-08-03), when calculating the score of (Donald Trump, Threaten with administrative sanction, Kim Jong-un, 2018-08-01), the second quadruplet visiting Switzerland does not contribute to the prediction of the relation between Donald Trump and Kim Jong-un since Switzerland is neutral. In such case, the model should find a way to skip the irrelevant snapshots based on the entity-related relation in the query. Therefore, an optimal TKGC model should consider (1) explicit temporal information and (2) implicit temporal information with skipping irrelevant snapshots by considering the query. 

In this paper, we propose Re-Temp, an innovative TKGC model designed for extrapolation settings that incorporates relation-aware temporal representation learning. The encoder of Re-Temp utilises explicit temporal embedding for each entity, combining static and dynamic embedding. Within the encoder, a sequential GNN is employed to capture the implicit temporal information with a skip information flow applied after each timestamp, taking into account the entity-related relation in the query. 
The main contributions of this paper can be summarised as follows:
\begin{itemize}
    \item We introduce Re-Temp, a precise TKGC model that leverages both explicit and implicit temporal information, incorporates a relation-aware skip information flow to exclude irrelevant information and adopts a two-phase forward propagation method to prevent information leakage\footnote{Code available at: https://github.com/adlnlp/re-temp}.
    \item We compare our Re-Temp against eight state-of-the-art baseline models from recent years using six publicly available TKGC datasets under the extrapolation setting. Our experimental results demonstrate that Re-Temp outperforms all of the baselines significantly.
    \item We conduct a detailed case study and statistical analysis to illustrate the distinct characteristics of each dataset and provide an explanation based on our experimental findings.
\end{itemize}

\begin{table*}[t!]
\caption{Summary of TKGC(extrapolation) models and our proposed model. The column`Temporal' presents the trend of the approach to how the temporal information is used, and the column `Query' shows the summary of the approach to how the model utilises query.}
\begin{adjustbox}{width=\linewidth}
\begin{tabular}{l|l|l|l}
\hline
\textbf{Method}   & \textbf{Core idea}                                               & \textbf{Temporal}                                 & \textbf{Query}                   \\\hline
RE-NET\cite{renet}   & estimate the future graph distribution                  & implicit & N/A                    \\
CyGNet\cite{cygnet}   & identify facts with repetition                          & explicit                                      & repetitive queries      \\
xERTE\cite{xerte}    & sample subgraph according to query                      & implicit & query-related subgraph\\
REGCN\cite{regcn}    & relation-GCN + GRU                                      & implicit & N/A                     \\
TANGO\cite{tango}    & neural ODE on continuous-time reasoning                 & implicit & N/A                     \\
TITER\cite{titer}    & path-based reinforcement learning                       & implicit  & query-related path    \\
CEN\cite{cen}      & ensemble model with different history lengths           & implicit  & N/A                 \\
HiSMatch\cite{hismatch} & two separated encoders for entity and query information & implicit & repetitive queries     \\\hline
Re-Temp (Ours)     &    skip irrelevant information according to entity-related relations                            & both   & query-related skip information flow  \\\hline     
\end{tabular}
\end{adjustbox}
\label{tab:related_work}
\end{table*}

\section{Related Work}
KGC models normally adopt an encoder-decoder framework\cite{DBLP:journals/debu/HamiltonYL17}, where the encoder generates the embedding of entities and relations and the score function plays as a decoder. Most of the existing works extend the static KGC models into TKGC models by introducing temporal information.


\subsection{TKGC(Interpolation)} 
To integrate the temporal information in the decoder, TTransE\cite{jiang2016towards} extends TransE\cite{bordes2013translating} with the summation of an extra timestamp embedding, and ConT\cite{ma2019embedding} extends Tucker\cite{balavzevic2019tucker} by replacing the learnable weight with the timestamp embedding. 
Some methods also focus on combining temporal information in the encoder: TA-DistMult\cite{garcia2018learning} encodes the temporal information into relation embedding by using LSTM, while DE-SimplE\cite{goel2020diachronic} encodes a diachronic entity embedding with temporal information. with decoders as DistMult and SimplE\cite{YangYHGD14a,kazemi2018simple} accordingly. These models produced relatively lower performance on TKGC under the extrapolation setting tasks since they are unable to capture unseen temporal information.

\subsection{TKGC(extrapolation)}
For the last few years, more attention has been paid to TKGC tasks under the extrapolation setting. GNNs are typically used as the encoder: RE-NET\cite{renet} applies sequential neighbourhood aggregators such as R-GCN\cite{schlichtkrull2018modeling} to get the distribution of the target timestamp snapshot, REGCN\cite{regcn} adopts CompGCN\cite{vashishth2020compositionbased} at each timestamp and GRU for sequential information. 
CEN\cite{cen} uses an ensemble model of sequential GNNs with different history lengths, TANGO\cite{tango} solves Neural Ordinary Equations and makes it as the input of a Multi-Relational GCN, and HiSMatch\cite{hismatch} builds two GNN encoders modelling the sequential candidate graph and query-related subgraphs separately and combines the representation from both sides into a matching function.   
Meanwhile, some methods do not follow the traditional encoder and decoder framework. xERTE\cite{xerte} extracts subgraph according to queries, CyGNet\cite{cygnet} identifies the candidates with repetition, and TITer\cite{titer} uses reinforcement learning methods to search for the temporal evidence chain for prediction.
To conclude, RE-NET, REGCN, and CEN adopt the entity evolvement information, while xERTE, CyGNet and TITer focus on the query. HiSMatch combines these two types of information with two separate encoders. However, none of the previous works encoded sequential and query-related information in one precise encoder. In addition to this, none of these methods considers explicit temporal information, except for CyGNet, which generates an independent timestamp vector but does not encode it into the entity or relation. Table \ref{tab:related_work} presents the summary of TKGC(extrapolation) models and emphasises the contribution of our proposed model.

\begin{figure*}
    \centering
    \includegraphics[width=0.95\linewidth]{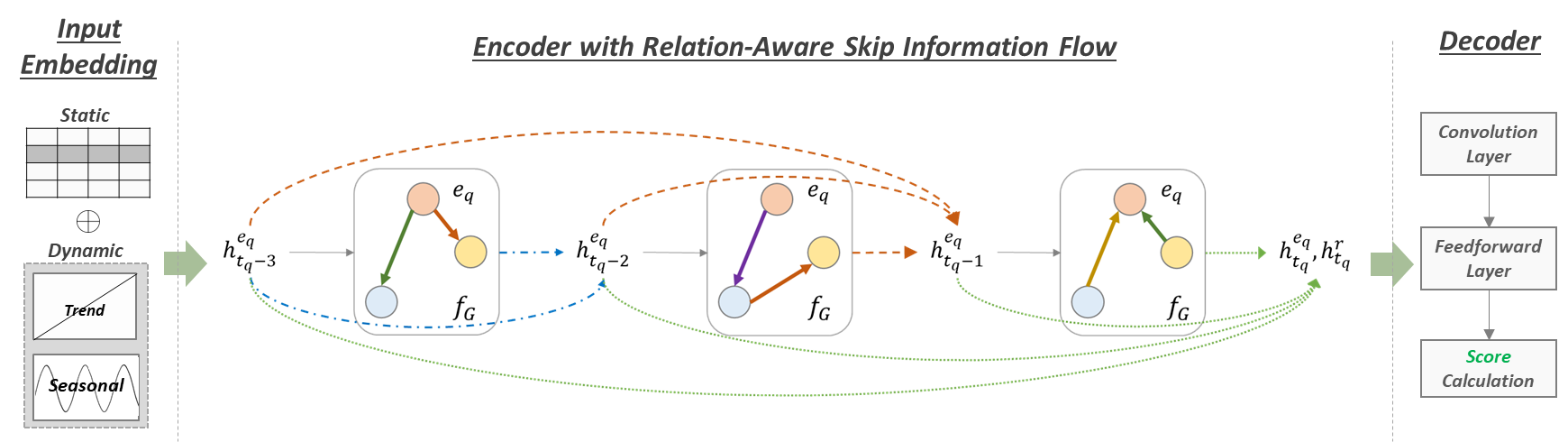}
    \caption{Illustration of Encoding and Decoding process in Re-Temp with history length as 3. For a query $q$, the input vector is $h_{t_q-3}^{e_q}$. The encoder with relation-aware skip information flow learns the entity and relation representation $h_{t_q}^{e_q}$ and $h_{r_q}$. Then the decoder measures the score of all the candidates.}
    \label{fig:model}
\end{figure*}

\section{Re-Temp}
The overall architecture of Re-Temp can be found in Figure \ref{fig:model}. Section \ref{sec:prob_form} describes the notations of a TKGC task. The input of the model is represented by a combination of static and dynamic entity embedding, in Section \ref{sec:input}, showing explicit temporal information. The encoder in Section \ref{sec:encoder} uses a sequential multi-relational GNN to learn implicit temporal information and after each timestamp, a relation-aware skip information flow mechanism is applied to retain the necessary information for prediction. The ConvTransE decoder together with the loss function is introduced in Section \ref{sec:decoder}. To avoid information leaking, we apply a two-phase forward propagation method in Section \ref{sec:avoid_info_leak}. 

\subsection{Problem Formulation}\label{sec:prob_form}
To denote the set of entities, relations, timestamps and facts, $\mathcal{E},\mathcal{R}, \mathcal{T}$ and $\mathcal{F}$ are selected. A temporal knowledge graph $G$ can be treated as $|\mathcal{T}|$ sequential snapshots, $G = \{G_0,G_1,...,G_T\}$, where $G_t = \{\mathcal{E},\mathcal{R}, \mathcal{F}_t\}$ is a directed multi-relational graph at timestamp $t$. For each fact, a quadruplet is represented as $(e_s, r, e_o, t)$, where $e_s, e_o \in \mathcal{E}$ are the subject and object entities, $r \in \mathcal{R}$ represents the relation and $t \in \mathcal{T}$ is the timestamp. The target of the temporal knowledge graph completion under the extrapolation setting is that for a query $q$, predicting $(e_s, r, ?, t_q)$ or $(?, r, e_o, t_q)$ given previous snapshots $\{G_0, G_1, ..., G_{t_q-1}\}$. Normally, the inverse of each quadruplet is added to the dataset, making all subject entity prediction problem $(?, r, e_o, t_q)$ into object entity prediction problem $(e_o, r^{-1}, ?, t_q)$.

\subsection{Explicit Temporal Representation}\label{sec:input}
For sequential snapshots with length $k$, let $h_{t_q-k}^{e_q} \in \mathbb{R}^{1\times d}$ denotes the input embedding of the subject entity $e_q$ from query $q$, and $d$ is the dimension of the input. 
In order to encode the explicit temporal information, we concatenated two kinds of input embedding; static and dynamic embedding. The static embedding reveals the nature of an entity that does not change through time, while the dynamic part reveals the time-dependent information. 

Inspired by ATiSE\cite{ATiSE}, the dynamic embedding is decomposed into the trend component and seasonal component, and the trend component can be represented as a linear transformation on $t$ while the seasonal component should be a periodical function of $t$. Thus, we model the dynamic temporal embedding at timestamp $t$ by the summation of trend embedding $w_{e_q,0}t$ and seasonal embedding $sin(2\pi w_{e_q,1}t)$. After concatenation with the static embedding, a feed-forward layer is applied. Formally, the input of the encoder $h_{t_q-k}^{e_q}$ is derived by:
\begin{equation}\label{eq:1}
    h_{t_q-k}^{e_q,S} = h^{e_q,S}
\end{equation}
\begin{equation}\label{eq:2}
    h_{t_q-k}^{e_q,D} = w_{e_q,0}(t_q-k) + sin(2\pi w_{e_q,1}(t_q-k))
\end{equation}
\begin{equation}\label{eq:3}
    h_{t_q-k}^{e_q} = W_{tmp}(h_{t_q-k}^{e_q,S}\oplus h_{t_q-k}^{e_q,D})
\end{equation}
where $h_{t_q-k}^{e_q,S}$ in Equation \ref{eq:1} and $h_{t_q-k}^{e_q,D}$ in Equation \ref{eq:2} denote the static and dynamic embedding for subject entity $e_q$ at timestamp $t_q-k$, $\oplus$ denotes the concatenation, and $h^{e_q,s}$, $w_{e_q,0}$, $w_{e_q,1}$, $W_{tmp}$ are learnable parameters. The major difference between our explicit temporal representation and ATiSE lies in the fact that employing a learnable feed-forward layer to concatenate the dynamic embedding and static embedding, enables the model to determine the extent to which it should utilise information from each embedding rather than simply utilising both.
Relation embedding $h^r$ can simply be extracted from a static embedding lookup table since we do not expect the relation's nature to evolve through time.

\newcolumntype{x}{>{\centering\arraybackslash}p{2.2cm}}

\begin{table*}[t!]
\centering
\caption{Statistics Details of Benchmark Dataset}
\begin{adjustbox}{width =1 \linewidth}
\begin{tabular}{l|x|x|x|x|x|x}
\hline
& \textbf{ICEWS14} & \textbf{ICEWS18} & \textbf{ICEWS05-15} & \textbf{ICEWS14*} &\textbf{ GDELT} & \textbf{WIKI}\\\hline
\# Entities & 7,128 & 23,033 & 10,094 & 7,128 & 7,691 & 12,554\\
\# Relations & 230 & 256 & 251 & 230 & 240 & 24\\
\# Facts & 89,730 & 468,558 & 461,329 & 90,730 & 2,277,405 & 669,934\\
\# Snapshots & 365 & 304 & 4,017 & 365 & 2,976 & 232\\
\# Snapshots in Train/Val/Test set & 304/30/31 & 240/30/34 & 3,243/404/370 & 262/52/51 & 2,304/288/384 & 211/11/10\\
\# Facts per Snapshot & 245.8 & 1541.3 & 32.2 & 248.6 & 765.3 & 2887.6\\
Time Interval & 1 day & 1 day & 1 day & 1 day & 15 mins & 1 year\\
Total Time Range & 1 year & 0.83 years & 11 years & 1 year & 0.54 years & 232 years \\\hline 
\end{tabular}
\end{adjustbox}
\label{tab:staitcs}
\end{table*}

\subsection{Relation-Aware Skip Information Flow}\label{sec:encoder}
In order to handle implicit temporal information, we use a sequential GNN-based encoder with a new relation-aware skip information flow mechanism. Following recent work\cite{regcn,cen,hismatch}, we adopt a variant of CompGCN\cite{vashishth2020compositionbased} at each timestamp to model the multi-relational snapshot, outputting the entity embedding $h_e$ and the relation embedding $h_r$. The details of CompGCN are shown in Appendix \ref{sec:appendix_compgcn}.

Not all snapshots in the recent history are useful in predicting query $q$, hence, a relation-aware skip information flow is applied. Two things are considered: (1) Skip connection is used for filtering out the unnecessary information from each timestamp. (2) Relation-aware attention mechanism helps to determine whether some information should be filtered. Thus, after getting the output of CompGCN, they will be weighted-summed up with previous timestamps input to partially skip the irrelevant snapshots. The weights of the weighted sum are calculated by considering both the entity and the entity-related relation in the query. 

Formally, for an entity $e_q$, the relation associated with $e_q$ should be considered. To capture the entity-related relation information, mean pooling is applied on all relation embedding associated with $e_q$ at timestamp $t_q$. The representation obtained from mean pooling will serve as a reference vector to help the model determine the information to keep or skip. Then, this average relation embedding will be summed with all $m$ previous timestamps one by one, followed by a feedforward layer. This calculation can also be treated as additive attention. After getting the attention weights $\beta_j^{e_q}$, the weighted sum using these attention weights is applied on the current CompGCN output $h_{t_i}^{e_q,L}$ and all $m$ previous timestamp inputs. The detailed calculation shows as follows:
\begin{equation}\label{eq:5}
    h_{r,t_q}^{e_q} = \frac{1}{|R_{t_q}^{e_q}|}\sum\limits_{r \in R_{t_q}^{e_q}}h_{r}
\end{equation}
\begin{equation}\label{eq:6}
    attn_j^{e_q} = \begin{cases}
    \mathbf{0}& j = 0\\
    W_{a}(h_{t_i-j}^{e_q} + h_{r,t_q}^{e_q})& j\in [1,m]\\
\end{cases}
\end{equation}
\begin{equation}\label{eq:7}
    \beta_j^{e_q} = softmax(attn_j^{e_q}), j\in [0,m]
\end{equation}
\begin{equation}\label{eq:8}
    h_{t_i+1}^{e_q} = \beta_0^{e_q}h_{t_i}^{e_q,L} + \sum\limits_{j=1}^m\beta_j^{e_q}h_{t_i-j}^{e_q}
\end{equation}
Note that the output of each timestamp is also the input of the next timestamp. Equation \ref{eq:5} shows the entity-associated relation embedding and $R_{t_q}^{e_q}$ denotes the relation set which connects with entity $e_q$ at timestamp $t_q$. Equation \ref{eq:6} and \ref{eq:7} denotes the attention score and weight calculation where $W_{a}$ is learnable. By applying the relation-aware skip information flow, our model is capable of skipping irrelevant snapshots by considering the target query relations.

\subsection{Decoder}\label{sec:decoder}
ConvTransE\cite{shang2019end} is widely used in both static KGC\cite{ malaviya2020commonsense} and TKGC\cite{hismatch} as the score function, and ours is no exception. After getting the score of each candidate using ConvTransE, we train the model as a classification problem and the loss function for each query shows as follows:
\begin{equation}
    L = - \sum\limits_{e_c \in \mathcal{E}}z_clog(s(e_q, r_q, e_c, t_q))
\end{equation}
and $z_c$ will be 1 if correctly classified, otherwise, it is 0. The training target is to minimise the total loss for all queries. Appendix \ref{sec:appendix_convtranse} introduces the details of ConvTransE.

\subsection{Two-Phase Propagation}\label{sec:avoid_info_leak}
There is a potential information leakage problem by applying the relation-aware information flow mechanism. Suppose a query in the test set is $(A, r, B, t)$, after adding the inverse of quadruplets, $(B, r^{-1}, A, t)$ will be in the test set. When applying the encoder, with the relation-aware skip information flow, $A$ and $B$ will contain the information of $r$ and $r^{-1}$ accordingly. Therefore, when making predictions on $(A, r, ?, t)$ and calculating the score by dot product $A$ and all candidates, there is a chance that the information of $r$ in $A$ can meet the information of $r^{-1}$ in $B$. Since $r$ and $r^{-1}$ are paired, the model might find a shortcut to determine $B$ is the right answer for $(A, r, ?, t)$. This information leakage will result in unreasonably high performance during evaluation.

To avoid such information leakage, we propose a two-phase forward propagation method. We divide the dataset into two subsets: the original set and the inverse set. The inverse set is the set of inverse quadruplets. The snapshot graph in the history will be built on the whole set, while during forward propagation, the original set and inverse set are used separately. The output of the original set and the inverse set will be collected for loss calculation or performance evaluation.

\section{Experiments}\label{tab:experiments}
\subsection{Experiment Setup}
\textbf{Datasets} We evaluated our model on six widely-used TKG datasets: \textit{ICEWS14}\cite{regcn}, \textit{ICEWS18}\cite{renet}, \textit{ICEWS05-15}\cite{xerte}, \textit{ICEWS14*}\cite{xerte}, \textit{GDELT}\cite{renet}, and \textit{WIKI}\cite{leblay2018deriving}. The overall statistics of each dataset are presented in Table \ref{tab:staitcs}. All datasets are split into the Training, Validation and Test sets in chronological order. For example, the timestamps in \textbf{ICEWS14} are from 1st to 304th, from 305th to 334th and from 335th to 365th for training, validation and test set accordingly.
\begin{itemize}
    \item \textbf{ICEWS14}, \textbf{ICEWS18}, \textbf{ICEWS05-15}, \textbf{ICEWS14*} are extracted from Integrated Crisis Early Warning System which is a database system recording political events. \textbf{14}, \textbf{18}, \textbf{05-15} represent the year of the dataset(2014, 2018, 2005-2015), and \textbf{ICEWS14*} uses a different split compared with \textbf{ICEWS14}. The time interval of \textbf{ICEWS} is 1 day. A sample from \textbf{ICEWS} datasets is (John\_Kerry, Host\_a\_visit, Benjamin\_Netanyahu, 2014-01-01) 
    \item \textbf{GDELT} is also a political event temporal knowledge graph dataset from the \textbf{G}lobal \textbf{D}atabase of \textbf{E}vents, \textbf{L}anguage, and \textbf{T}one\cite{leetaru2013gdelt}. Compared with \textbf{ICEWS} datasets, its time interval is only 15 minutes and \textbf{GDELT} is collected from a wider variety of sources. (Minist, Return, Nigeria, 0) is a sample in \textbf{GDELT}.
    \item \textbf{WIKI} is from Wikidata, an open knowledge base and not limited to political events. The temporal representation in the facts from Wikidata is not a single date/year but a range. For example, the fact (Wang Shu, educated at, Southeast University) is valid from 1981 to 1988. To represent a such range, \textbf{WIKI} generates eight quadruplets across eight snapshots during 1981-1988.
\end{itemize}
All the datasets are consistent with their intended use.

\textbf{Baselines} Our Re-Temp is compared with TKGC models under the extrapolation setting. Eight models from recent years are selected as baselines: \textbf{RE-NET}\cite{renet}, \textit{RE-GCN}\cite{regcn}, \textit{CyGNet}\cite{cygnet}, \textit{xERTE}\cite{xerte}, \textit{TITer}\cite{titer}, \textit{TANGO}\cite{tango}, \textit{CEN}\cite{cen}, and \textit{HiSMatch}\cite{hismatch}. Models that are designed for static KG completion or TKGC under the interpolation setting tasks are not compared since they naturally perform badly in TKGC under the extrapolation setting tasks. 

\begin{table*}[t!]
\caption{Performance(\%) with Baseline models. The highest value is bold and the second highest is underlined.}
\centering
\begin{adjustbox}{width=1\textwidth}
\begin{tabular}{l|cccc|cccc|cccc}
\hline
\multirow{2}{*}{\textbf{Model}} & \multicolumn{4}{c|}{\textbf{ICEWS14}}       & \multicolumn{4}{c|}{\textbf{ICEWS18}}    & \multicolumn{4}{c}{\textbf{ICEWS05-15}}    \\\cline{2-13}
& \textbf{MRR}  & \textbf{hits@1} & \textbf{hits@3} & \textbf{hits@10} & \textbf{MRR}  & \textbf{hits@1} & \textbf{hits@3} & \textbf{hits@10}  & \textbf{MRR}  & \textbf{hits@1} & \textbf{hits@3} & \textbf{hits@10}  \\\hline
RE-NET\cite{renet}   & 37.01 & 27.02 & 39.66 & 54.85 & 29.02 & 20.03 & 33.14 & 48.60  & 44.03 & 34.43 & 49.03 & 64.03 \\
CyGNet\cite{cygnet}  & 35.02 & 25.72 & 39.06 & 53.50  & 25.03 & 16.03 & 29.28 & 43.42 & 37.03 & 27.01 & 42.23 & 56.98 \\
xERTE\cite{xerte}    & 40.12 & 32.11 & 44.73 & 56.25 & 29.31 & 21.03 & 33.51 & 46.48 & 46.62 & 37.84 & 52.31 & 63.92 \\
REGCN\cite{regcn}     & 41.50  & 30.86 & 46.60  & 62.47 & 30.55 & 20.00 & 34.73 & 51.46 & 46.41 & 35.17 & 52.76 & 67.64 \\
TANGO\cite{tango}     & 30.12 & 23.03 & 35.48 & 52.32 & 28.97 & 19.51 & 32.61 & 47.51 & 42.86 & 32.72 & 48.14 & 62.34 \\
TITer\cite{titer}      & 41.73 & 32.74 & 46.46 & 58.44 & 29.96 & 22.06 & 33.41 & 44.92 & 47.78 & 38.05 & 53.11 & 65.93 \\
CEN\cite{cen}        & 42.20  & 32.08 & 47.46 & 61.31 & 31.50  & 21.70  & 35.44 & 50.59 & 45.97 & 35.56 & 51.45 & 66.14 \\
HiSMatch\cite{hismatch}   & \underline{46.42} & \underline{35.91} & \underline{51.63} & \underline{66.84} & \underline{33.99} & \underline{23.91} & \underline{37.90}  & \underline{53.94} & \underline{52.85} & \underline{42.01} & \underline{59.05} & \underline{73.28} \\\hline
Re-Temp (Ours) & \textbf{48.04} & \textbf{37.32} & \textbf{53.60} & \textbf{68.90} & \textbf{35.82} & \textbf{25.02} & \textbf{40.36} & \textbf{57.30} & \textbf{56.30} & \textbf{45.49} & \textbf{62.80} & \textbf{77.17}\\\hline 

\multicolumn{1}{l}{}\\

\hline
\multirow{2}{*}{\textbf{Model}} & \multicolumn{4}{c|}{\textbf{ICEWS14*}}       & \multicolumn{4}{c|}{\textbf{GDELT}}    & \multicolumn{4}{c}{\textbf{WIKI}}    \\\cline{2-13}
& \textbf{MRR}  & \textbf{hits@1} & \textbf{hits@3} & \textbf{hits@10} & \textbf{MRR}  & \textbf{hits@1} & \textbf{hits@3} & \textbf{hits@10}  & \textbf{MRR}  & \textbf{hits@1} & \textbf{hits@3} & \textbf{hits@10}  \\\hline
RE-NET\cite{renet}   & 38.28 & 28.68 & 41.43 & 54.52 & 19.63 & 12.39 & 21.03 & 34.02 & 49.66 & 46.98 & 51.23 & 53.49 \\
CyGNet\cite{cygnet}   & 33.13 & 24.16 & 37.02 & 51.23 & 18.98 & 12.32 & 20.56 & 33.89 & 43.78 & 39.02 & 46.12 & 51.92 \\
xERTE\cite{xerte}   & 40.77 & 32.65 & 45.71 & 57.29 & 18.07 & 12.31 & 20.05 & 30.32 & 71.16 & 68.03 & 76.15 & 78.99 \\
REGCN\cite{regcn}  & 41.79 & 31.55 & 46.67 & 61.53 & 19.31 & 11.99 & 20.61 & 33.59 & 77.58 & 73.72 & 80.39 & 83.69 \\
TANGO\cite{tango}    & 26.35 & 17.33 & 29.27 & 44.32 & 18.03 & 12.36 & 19.96 & 29.31 & 51.15 & 49.65 & 52.26 & 53.44 \\
TITer\cite{titer} & 41.76 & 32.69 & 46.35 & 58.46 & 17.02 & 11.23 & 19.81 & 26.92 & 75.51 & 72.98 & 77.51 & 79.32 \\
CEN\cite{cen}      & 40.78 & 31.26 & 45.26 & 59.16 & 19.89 & 12.61 & 21.16 & 34.09 & 77.65 & 73.86 & 80.69 & 84.00    \\
HiSMatch\cite{hismatch} & \underline{45.82} & \underline{35.84} & \underline{50.79} & \underline{65.08} & \underline{22.01} & \underline{14.45} & \underline{23.80}  & \underline{36.61} & \underline{78.07} & \underline{73.89} & \underline{81.32} & \textbf{84.65} \\\hline
Re-Temp (Ours)     & \textbf{46.40} & \textbf{35.86} & \textbf{51.69} & \textbf{67.12} & \textbf{25.05} & \textbf{15.70} & \textbf{27.14} & \textbf{44.16} & \textbf{78.51} & \textbf{74.80} & \textbf{81.33} & \underline{84.50} \\\hline 
\end{tabular}
\end{adjustbox}
\label{tab:basline2}
\end{table*}

\newcolumntype{y}{>{\arraybackslash}p{3.2cm}}
\begin{table}[t!]
\caption{Cases from WIKI Dataset about Lionel Messi from Year 2003 to Year 2005.}
\label{tab:wiki_case}
\begin{adjustbox}{width=1\linewidth}
    \begin{tabular}{l|l|y|l}\hline
        \textbf{Subject Entity} & \textbf{Relation} & \textbf{Object Entity} & \textbf{Year} \\\hline
         Lionel Messi & residence & Barcelona &  2003 \\
         Lionel Messi & member of sports team & FC Barcelona C & 2003 \\\hline
         Lionel Messi & residence & Barcelona &  2004 \\
         Lionel Messi & member of sports team & FC Barcelona C & 2004\\
         Lionel Messi & member of sports team & FC Barcelona Atlètic & 2004\\\hline
        Lionel Messi & residence & Barcelona & 2005\\
        Lionel Messi & member of sports team & Argentina national football team & 2005\\\hline
    \end{tabular}
\end{adjustbox}
\end{table}

\textbf{Hyperparameter} Following the previous works\cite{cen,hismatch}, the dimension of the input is set to 200, which is also the hidden dimension of the graph model and decoder hidden dimension. The number of graph neural network layers is 2 and the dropout rate is set to 0.2. Adam\cite{DBLP:journals/corr/KingmaB14} with a learning rate of 1e-3 is used for optimisation. The model is trained on the training set with a maximum of 30 epochs and we stop training when the validation performance doesn't improve in 5 consecutive epochs. Then, the test set is evaluated using the trained model.

\textbf{Evaluation Metrics} Following the previous works\cite{xerte,cygnet,hismatch}, we employ widely used evaluation metrics, Mean Reciprocal Rank(MRR), hits@1, hits@3, and hits@10, which is explained in Appendix \ref{sec:appendix_evaluation_metrics}. We adopt the way of filtering out the quadruplets occurring at the query time, followed by \citet{titer,tango}, and we report the five-times running average result.

\subsection{Performance Comparison}\label{sec:basline}
We use a history length of 3 for ICES14, ICEWS18, ICEWS05-15, ICEWS14* and GDELT, while 1 for WIKI. The influence of history length is discussed in Section \ref{sec:history}. Table \ref{tab:basline2} presents the performance comparison of all baseline models. Our model, Re-Temp, outperforms significantly almost all the baseline models on all datasets, indicating the superiority of our Re-Temp model. In detail, three points can be observed:

Firstly, HiSMatch\cite{hismatch} achieved the second-highest performance on most of the datasets by considering both the query subgraph and entity subgraph. The concept considering both query and entity of HiSMatch is similar to our relation-aware attention mechanism in the skip information flow. However, HiSMatch only builds the query subgraph using the exact same relation of the query, which ignores the potential similarity between relations. For example, in ICEWS14, when making a prediction on (A, provide\_aid, ?, $t_q$), relation `provide\_aid' and `provide\_military\_aid' share similarities, but HisMatch only considers the entity with `provide\_military\_aid' in the recent history while our method uses the embedding of relation to calculate the attention weights, making it general for different types of relations that are close in the embedding space and outperforming HiSMatch. Meanwhile, HiSMatch builds two separate encoders and fuses the output for the decoder while our model only applies one encoder for better information alignment.

Secondly, among four ICEWS datasets, our model achieves more improvement on ICEWS05-15. As shown in Table \ref{tab:staitcs}, the snapshots in ICEWS05-15 are sparser than others, showing the ability of our model to learn sequential information with less data.

Thirdly, our model only achieves a comparable performance with HiSMatch on WIKI, which might result from the nature of this dataset. Table \ref{tab:wiki_case} lists some cases of facts about Lionel Messi in WIKI. Suppose giving the quadruplets from 2003 and 2004, it is relatively easy to predict (Lionel Messi, residence, ?, 2005) based on his previous residence, however, it is almost impossible to have a correct prediction on (Lionel Messi, residence, ? , 2005) since the previous snapshots don't provide enough information on Argentina national football team. This is an issue in WIKI: the predictions are either too easy (using the previous facts), or too difficult (even humans can not make a correct prediction without any external knowledge). Thus, a relatively better model is not enough to generate an undoubtful better performance on WIKI, and our model and some previous baseline models (CEN, HiSMatch) share similar results on this dataset.

\begin{figure}[t!]
    \centering
    \includegraphics[width=0.95\linewidth]{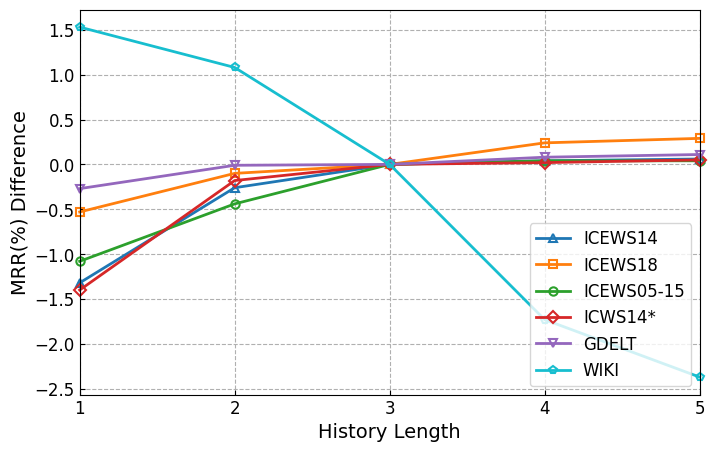}
    \caption{MRR(\%) change of Re-Temp with the history lengths. The x-axis is the history length and the y-axis is the MRR(\%) change compared with history length 3.}
    \label{fig:history}
\end{figure}

\begin{figure}[t!]
    \centering
    \includegraphics[width=0.95\linewidth]{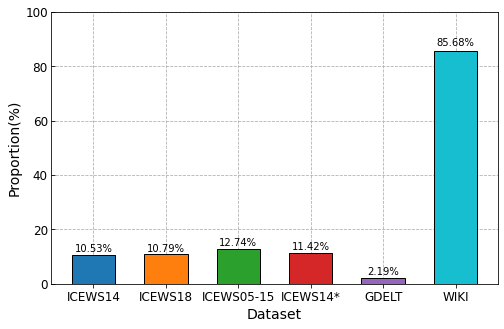}
    \caption{Proportion(\%) of quadruplets shown in exact one timestamp before for each dataset. The x-axis is the name of the dataset and the y-axis is the proportion(\%).}
    \label{fig:proportion}
\end{figure}

\begin{table*}
\caption{The MRR(\%) result of the ablation test of Re-Temp. The highest value is bold.}
\centering
\begin{adjustbox}{width = 0.8\linewidth}
\begin{tabular}{l|x|x|x|x|x|x}
\hline
\textbf{Model} & \textbf{ICEWS14} & \textbf{ICEWS18} & \textbf{ICEWS05-15} & \textbf{ICEWS14*} & \textbf{GDELT} & \textbf{WIKI}  \\\hline
 Re-Temp & \textbf{48.04} & \textbf{35.82}  & \textbf{56.30} & \textbf{46.40} & \textbf{25.05} & 78.51\\
 - \textit{dynamic} & 47.52   & 35.33  & 55.12 & 45.89 &  24.85 & 76.04 \\
 - \textit{relation\_aware} & 39.93  & 30.56 &  44.95 & 38.75 & 19.92 & 78.14\\
 - \textit{skip} & 36.56   &  28.07  & 43.80 & 36.30 & 18.61 & \textbf{79.60}\\\hline
\end{tabular}
\end{adjustbox}
\label{tab:ablation}
\end{table*}

\subsection{Impact of history length}\label{sec:history}
To study the impact of history length on different datasets, experiments with different history lengths are conducted. The default value of history length is 3 and the MRR changes in percentage are shown in Figure \ref{fig:history} with history lengths from 1 to 5. Two major points can be noticed:

(1) On most of the datasets (ICEWS14, ICEWS18, ICEWS05-15, ICEWS14*, and GDELT), a larger history length results in a higher MRR. Where the history length is small, enlarging the history length can substantially enhance performance. However, when the history length surpasses three, the degree of improvement becomes marginal. This aligns with the expectations that the recent several snapshots can help with inference, while in a long history, the irrelevant information does not contribute to the performance. By considering the model performance and calculation complexity, history length = 3 is selected as the final model for these datasets.

(2) An exception occurs on WIKI, where the model achieves the best performance when history length = 1. To investigate the factors, a detailed statistical analysis of the datasets is conducted. Table \ref{tab:wiki_case} in Section \ref{sec:basline} shows some sample queries in WIKI, where some facts are the same as the facts at previous timestamps, the reason lies in that for a fact ($s$,$r$,$o$,$t_1$ - $t_n$), WIKI generates the same quadruplets across the time range from $t_1$ to $t_n$. Figure \ref{fig:proportion} shows the proportion of the quadruplets at $t_q$ shown in the previous timestamp $t_q-1$ for all timestamps in the test set on each dataset. 85.68\% samples in the WIKI show in the one timestamp before, while fewer than 15\% samples in ICEWS14, ICEWS18, ICEWS05-15, ICEWS14*, GDELT are from the previous timestamp. The same quadruplets shown across different timestamps in WIKI result in similar snapshots(graphs) at different timestamps. When a larger history length is applied, multiple graph neural network models applied on multiple similar graphs will be approximated to applying a multiple layers GNN model on one graph, which leads to the over-smoothing issue in a deep GNN\cite{li2018deeper}. Therefore, a large history length may decrease model performance on WIKI.

\begin{table*}
\caption{MRR(\%) of our model with different ensemble methods. The highest value is bold.}
\centering
\begin{adjustbox}{width=0.8\linewidth}
\begin{tabular}{l|x|x|x|x|x}
\hline
\textbf{Ensemble Model} & \textbf{ICEWS14} & \textbf{ICEWS18} & \textbf{ICEWS05-15} & \textbf{ICEWS14*} & \textbf{GDELT}   \\\hline
 Re-Temp & 48.04 & 35.82 & 56.30 & 46.40  &  25.05 \\
Ensemble (avg pooling)& 48.58 & 36.16  & \textbf{56.72} & 46.56 & 25.04  \\
 Ensemble (max pooling)& \textbf{48.69} & \textbf{36.38}  & 56.69 & \textbf{47.06} & \textbf{25.06}  \\
Ensemble (min pooling)& 47.55 & 35.72  & 55.58 & 46.23 & 25.03  \\\hline
\end{tabular}
\end{adjustbox}
\label{tab:ensemble}
\end{table*}

\subsection{Ablation Study}\label{sec:ablation}
Table \ref{tab:ablation} presents the ablation study of different components of our model. 

\textbf{Impact of explicit temporal embedding} To evaluate the efficiency of the \textbf{explicit temporal representation}, we remove the dynamic embedding from the explicit temporal input, resulting in only the static embedding of each entity left. For all six benchmark datasets, removing dynamic embedding leads to worse performance. Compared with the performance drop in ICEWS14, ICEWS18, ICEWS14* and GDELT, it is clear that the MRR decreases more in WIKI and ICEWS05-15. The reason is that the total time range in these two datasets is large (232 years and 11 years), and the entity information can evolve over a long period, which can be captured by explicit temporal embedding.

\textbf{Impact of relation-aware skip information flow} To demonstrate how the relation-aware skip information flow contributes to the model performance, two ablation tests are conducted. (1)`-relation\_aware' means that when calculating the attention score in skip information flow, the entity-related relation is omitted, formally, the attention score is Equation \ref{eq:6} is changed to:$attn_j^{e_q} = W_{a}(h_{t_i-j}^{e_q}), j\in [1,m]$. (2)`-skip' means removing the whole skip information flow, making the input of each timestamp the last timestamp the output: $h_{t_i+1}^{e_q} = h_{t_i}^{e_q,L}$.

The model performance drops heavily if no relation-aware attention mechanism is applied, showing the vital importance of the relation-aware attention mechanism. We can conclude that the entity-related relation information actually helps the model to select necessary information. In most cases, removing the skip connection worsens the model performance compared with only removing the relation-aware attention mechanism. Compared with `-relation\_aware' setting, the models under the `-skip' setting learn from all the recent snapshots for prediction, leading to the involvement of irrelevant information during prediction. 

However, WIKI shows better performance under this setting, even compared with our original Re-Temp model. The reason might be the same as that discussed in Section \ref{sec:history}: More than 80\% of facts in the WIKI show in the previous timestamp, and a graph model applied on the previous timestamp can easily capture that repetitive information for prediction.

\subsection{Ensemble Modelling Evaluation}\label{sec:ensemble}
CEN\cite{cen} builds an ensemble model with different history lengths. Inspired by this, we test our model under an ensemble setting. For a model with a history length of $k$, suppose the score vector of all candidates for query $q$ is $s_k^q$, a pooling method is applied on $\{s_1^q, s_2^q, ..., s_k^q\}$ to get the final score. Three different pooling methods are applied. Table \ref{tab:ensemble} shows the MRR(\%) results of our model under the ensemble setting. We applied the history lengths from one (1), and the maximum history length is set to three (3) as previously defined. We did not include the experiments on WIKI since the optimal history length is one (1), and no models with smaller history lengths can be used. 
First of all, our model can benefit under the ensemble setting on four of the datasets (ICEWS14, ICEWS18, ICEWS05-15, ICEWS14*), but only achieve similar performance on GDELT compared with the original Re-Temp model (25.05\%). Considering the history length influence shown in Figure \ref{fig:history}, the model achieves similar results with different history lengths. Therefore, models with different history lengths on GDELT might be similar making the ensemble models less effective. However, ICEWS datasets are history-length sensitive, and ensemble models can benefit from different models of different history lengths. In addition to this, max pooling usually achieves the best performance as the ensemble method while min pooling will worsen the performance.

\section{Conclusion}
We introduced Re-Temp, which integrates both explicit and implicit temporal information and applies a relation-aware skip information flow to adopt after each timestamp to remove unnecessary information for prediction by taking the entity-related relation in the query into consideration. The experimental results on six TKGC datasets present the superiority of our model, compared with eight baseline models. We also conduct a statistical analysis of the datasets to show the different nature between WIKI and other datasets. It is hoped that Re-temp presents insight into the importance of the relation in the query and both types of temporal information.

\section{Limitations}
Re-Temp still follows Knowledge Graph Completion encoder-decoder framework\cite{DBLP:journals/debu/HamiltonYL17} while more frameworks can be explored. The graph model at each timestamp and the decoder score function follow the same methods widely used by other models. 

Since we have shown that the explicit temporal embedding and the skip information flow contribute to model performance, more work can be done by combining these concepts into the graph model and score function, for example, combining the entity-related relation into the graph model at each timestamp to selectively propagate between nodes, or combining the explicit temporal embedding into the decoder score function. Also, like most TKGC models, Re-Temp can not handle new entities that do not show in the training data. More methods integrating the text description can be explored\cite{lv2022pre}.

\bibliography{anthology,custom}
\bibliographystyle{acl_natbib}

\appendix
\section{Model Component Details}
\subsection{CompGCN}\label{sec:appendix_compgcn} 
In CompGCN, at each layer, edges(relations) are conducted as the transformation on the connected node(entity), and then a weighted sum calculation from GCN\cite{kipf2017semi} is applied to the transformed entity. Self-loop is also calculated before the activation function. Formally, for a entity node $e_q$ at timestamp $t_i$ at $l$th layer, the propagation shows as follows:
\begin{equation}
    h_{t_i}^{e_q,l+1} = \sigma(\frac{1}{|N_{t_i}^{e_q}|}\sum\limits_{e_n \in N_{t_i}^{e_q}}W_{g,0}^lf(h_{t_i}^{e_n,l},h_r) 
    + W_{g,1}^lh_{t_i}^{e_q,l})
\end{equation}
where $N_{t_i}^{e_q}$ is the set of the neighbour entities of $e_q$ at timestamp $t_i$, $\sigma$ is the activation function and RReLU\cite{xu2015empirical} is chosen. $W_{g,0}^l$ and $W_{g,1}^l$ are learnable parameters at layer $l$, and $f$ is the composition function for neighbour entity embedding $h_{t_i}^{e_n,l}$ and relation embedding $h_r$, such as summation, subtraction, element-wise product, or circular-correlation\cite{xu2015empirical}.Summation is selected for better alignment of relation-aware skip information flow.

\begin{table*}
\caption{Re-Temp running time and number of parameters}
\centering
\begin{adjustbox}{width=0.9\linewidth}
\begin{tabular}{c|l|x|x|x|x|x|x}
\hline
\multicolumn{2}{c|}{} & \textbf{ICEWS14} & \textbf{ICEWS18} & \textbf{ICEWS05-15} & \textbf{ICEWS14*} & \textbf{GDELT} & \textbf{WIKI}   \\\hline
\multirow{2}{*}{\makecell{Running\\Time (min)}} & Training & 11.6 & 22.1 & 99.5 & 8.8 & 112.6 & 8.1 \\ \cline{2-8}
~ & Inference & 0.05 & 0.1 & 0.6 & 0.1 & 1.1 & 0.2  \\ \hline
\multirow{4}{*}{\makecell{Number of\\Parameters}}
 & Input & 4.4M & 14.0M & 6.2M & 4.4M & 4.8M & 7.6M \\ \cline{2-8}
~ & Encoder & 0.1M & 0.1M & 0.1M & 0.1M & 0.1M & 0.1M \\ \cline{2-8}
~ & Decoder & 2M & 2M & 2M & 2M & 2M & 2M \\ \cline{2-8}
~ & Total & 6.6M & 16.1M & 8.4M & 6.6M & 6.9M & 9.7M  \\ \hline
\end{tabular}
\end{adjustbox}
\label{tab:running_time}
\end{table*}

\begin{table*}
\caption{MRR(\%) result of the Encoder and Decoder Variants test. The highest value is bold.}
\centering
\begin{adjustbox}{width=0.9\linewidth}
\begin{tabular}{l|x|x|x|x|x|x}
\hline
\textbf{Model Variants} & \textbf{ICEWS14} & \textbf{ICEWS18} & \textbf{ICEWS05-15} & \textbf{ICEWS14*} & \textbf{GDELT} & \textbf{WIKI}  \\\hline
 Default & \textbf{48.04} & \textbf{35.82}  & \textbf{56.30} & \textbf{46.40} & \textbf{25.05} & \textbf{78.51}\\\hline
CompGCN (Element-Wise) & 47.57 & 35.24 & 55.81 & 45.54 & 24.98 & 70.99 \\
CompGCN (Circle-Correlation) & 46.69 & 35.09 & 56.00 & 44.65 & 24.90 & 74.32 \\\hline
Tucker & 46.36 & 35.14 & 56.84 & 44.48 & 24.65 & 78.28\\
DistMult & 34.48 & 22.85 & 39.80 & 36.58 & 18.18 & 59.35\\\hline
\end{tabular}
\end{adjustbox}
\label{tab:model_variant}
\end{table*}

\subsection{ConvTransE}\label{sec:appendix_convtranse}
By applying ConvTransE, the query subject entity embedding $h_{t_q}^{e_q}$ and query relation embedding $h_{r_q}$ are concatenated first, and then a convolutional layer and a feed-forward layer are applied. The score of each candidate is the dot-product of the candidate entity embedding with the representation after the ConvTransE. To denote the process of calculating the score of the candidate entity $e_c$:
\begin{equation}
    s(e_q, r_q, e_c, t_q) = h_{t_q}^{e_c}\text{FC}(\text{Conv1d}([h_{t_q}^{e_q}\oplus h_{r_q}]))
\end{equation}
where $e_c$ is the candidate entity.

\section{Experiment Setup Details}

\subsection{Running Details}\label{sec:appendix_hp}
All the models are trained by using 16 Intel(R) Core(TM) i9-9900X CPU @ 3.50GHz and NVIDIA Tesla P100 PCIe 16 GB. 

The number of parameters of Re-Temp can be decomposed into three parts:
\begin{itemize}
    \item \textbf{Input} Entity embedding: $3d|\mathcal{E}|+2d^2$, Relation embedding: $2d|\mathcal{R}|$
    \item \textbf{Encoder} CompGCN: $2d^2$, Relation-aware information flow: $d^2$
    \item \textbf{Decoder} ConvTransE: $ch(2ke+d+2)$, where $ch$ is the number of channels and $ke$ is the kernal size.
\end{itemize}
The running time and number of parameters of Re-Temp on different datasets under the default hyperparameters can be found in Table \ref{tab:running_time}.

\subsection{Evaluation Metrics}\label{sec:appendix_evaluation_metrics}
For each query, the model produces a ranked list of all possible candidates and the reciprocal rank is the inverse of the rank position of the correct answer. MRR is calculated by $\frac{1}{Q}\sum_{q=1}^Q\frac{1}{rank_q}$, which is the average reciprocal rank of all queries. Hits@N measures the proportion of results, where the correct answer is in the top $N$ ranked results. $N = 1,3,10$ are chosen, as all previous works adopted. The higher value of MRR and hits@N indicates the better performance of a model.

\section{Model Varirants Experiments}\label{sec:model_var}
We adopted CompGCN as a graph model in the encoder to model the multi-relational snapshot, and the transformation function is the sum: $f(h_{t_i}^{e_n,l},h_r) = h_{t_i}^{e_n,l} + h_r$. Followed by \citet{vashishth2020compositionbased}, we tested the default setting with the element-wise product or circle-correlation as the transformation function, as shown in Table \ref{tab:model_variant}. Even though good performance can be achieved by replacing the summation with other transformation functions, the summation is the best transformation function. The reason would be that during the skip information flow, additive attention is applied, which can benefit from the alignment of the entity embedding and relation embedding. Moreover, various decoders aside from ConvTransE are also experimented followed by TANGO\cite{tango}. As a decoder, Tucker\cite{balavzevic2019tucker} achieves much better performance than DistMult\cite{YangYHGD14a}. This is because DistMult lacks learnable parameters, while the learnable parameters in ConvTransE and Tucker give the model more complexity to have more possibility to find an optimal solution.

\section{Responsible Research - Risk}
Most temporal knowledge graph datasets focus on political news, which might raise concerns when predicting future political events where people have different political leanings.

\end{document}